\documentclass[letterpaper, 10 pt, conference]{ieeeconf}
\IEEEoverridecommandlockouts

\usepackage[utf8]{inputenc} 
\usepackage[T1]{fontenc}
\usepackage{amsmath}
\usepackage{amsfonts}
\usepackage{float}
\usepackage{graphicx}
\usepackage{booktabs}
\usepackage[font=small]{caption}
\usepackage{subcaption}
\usepackage[ruled]{algorithm2e}
\usepackage{booktabs} 
\usepackage{enumerate}
\usepackage{times}
\usepackage{soul}
\usepackage{url}
\usepackage{hyperref}
\usepackage{bm}
\usepackage{adjustbox}
\usepackage{xcolor}
\usepackage{placeins}
\usepackage[utf8]{inputenc}
\usepackage{booktabs}
\usepackage{array, multirow}
\usepackage{cleveref}
\usepackage{tikz}
\usepackage{wrapfig}
\usepackage{adjustbox}
\usepackage{multirow}
\usepackage{url}
  
\begin{document}

\newcommand\mycommfont[1]{\footnotesize\rmfamily\textcolor{blue}{#1}}

\setlength{\fboxrule}{1pt}
\setlength{\fboxsep}{0pt}  

\setlength{\textfloatsep}{2pt}
\setlength{\dbltextfloatsep}{2pt}  
\setlength{\dblfloatsep}{2pt}      
\setlength{\abovecaptionskip}{2pt} 
\setlength{\belowcaptionskip}{2pt}

\title{\LARGE \bf ConEQsA: Concurrent and Asynchronous \\ Embodied Questions Scheduling and Answering}

\author{Haisheng Wang$^{1}$ \and Dong Liu$^{2}$ \and Weiming Zhi$^{3,4}$
\thanks{email: \url{10235101559@stu.ecnu.edu.cn}; \url{dong.liu@aya.yale.edu}; \url{ weiming.zhi@sydney.edu.au}.}%
\thanks{$^{1}$ Software Engineering Institute, East China Normal University, Shanghai, China}%
\thanks{$^{2}$ Department of Computer Science, Yale University, CT, USA.}%
\thanks{$^{3}$ School of Computer Science, The University of Sydney, Australia.}%
\thanks{$^{4}$ College of Connected Computing, Vanderbilt University, TN, USA.}%
}

\maketitle

\begin{figure*}[h]
    \centering
    \includegraphics[width=\textwidth]{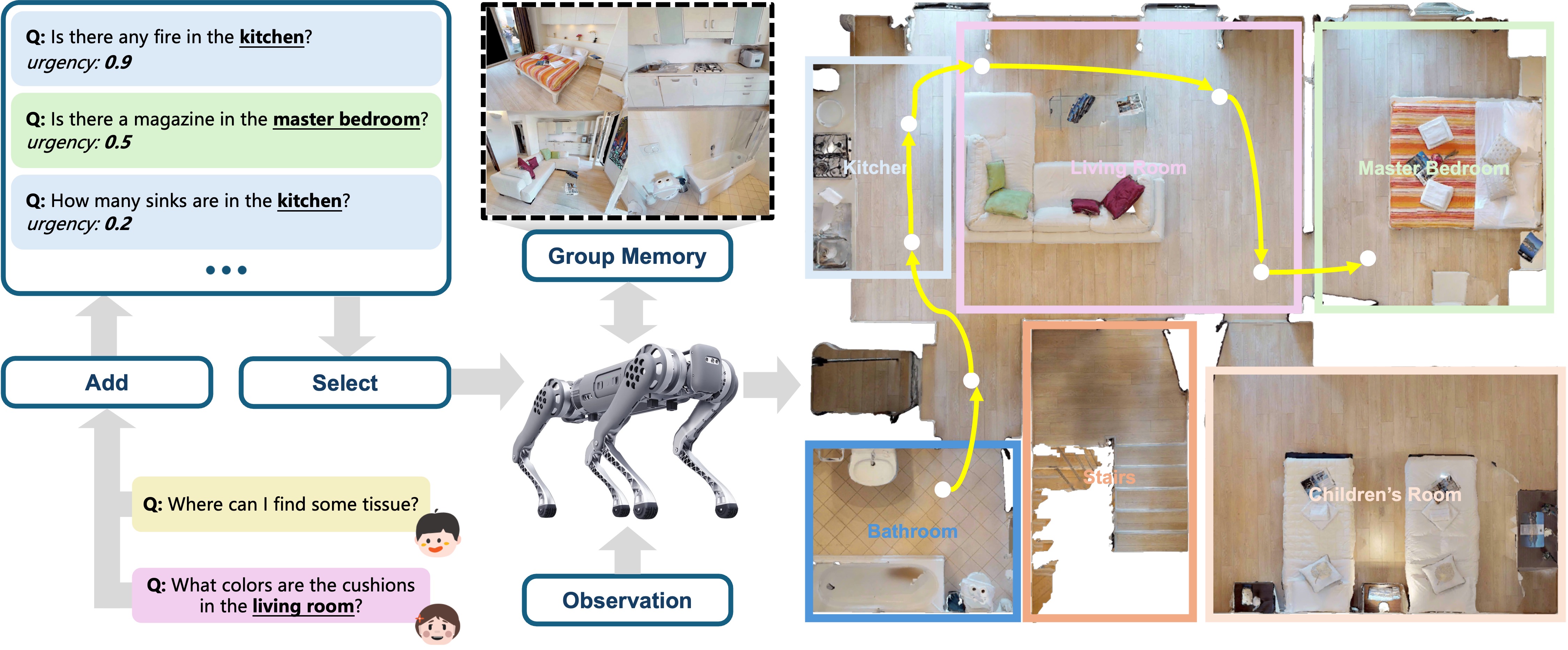}
    \caption{An overview of our Concurrent Embodied Questions Answering (ConEQsA) framework. Both initial questions and asynchronous follow-up questions with various urgency levels are coordinated via shared group memory and priority planning. }
    \label{fig:figure_1}
\end{figure*}

\begin{abstract}
This paper formulates the Embodied Questions Answering (EQsA) problem, introduces a corresponding benchmark, and proposes an agentic system to tackle the problem. Classical Embodied Question Answering (EQA) is typically formulated as answering \emph{one single} question by actively exploring a 3D environment. Real deployments, however, often demand handling \emph{multiple} questions that may arrive asynchronously and carry different urgencies. We formalize this setting as \textbf{E}mbodied \textbf{Q}uestion\textbf{s} \textbf{A}nswering (EQsA) and present \textbf{ConEQsA}, an agentic framework for \emph{concurrent, urgency-aware} scheduling and answering. ConEQsA leverages shared group memory to reduce redundant exploration, and a priority-planning method to dynamically schedule questions. To evaluate the EQsA setting fairly, we contribute the \textbf{C}oncurrent \textbf{A}synchronous \textbf{E}mbodied \textbf{Q}uestion\textbf{s} (CAEQs) benchmark containing 40 indoor scenes and five questions per scene (200 in total), featuring asynchronous follow-up questions and \emph{human-annotated} urgency labels. We further propose metrics for EQsA performance: Direct Answer Rate (DAR), and Normalized Urgency-Weighted Latency (NUWL), which serve as a fair evaluation protocol for EQsA. Empirical evaluations demonstrate that ConEQsA consistently outperforms strong \emph{sequential} baselines, and show that urgency-aware, concurrent scheduling is key to making embodied agents responsive and efficient under realistic, multi-question workloads. Code is available on \url{https://anonymous.4open.science/r/ConEQsA}.
\end{abstract}

\section{Introduction}\label{sec:Introduction}

Embodied agents, such as household assistants and service robots, are increasingly expected to operate effectively in complex, human-centric environments. Embodied Question Answering (EQA) is a fundamental precursor for integrating these agents into complex and interactive environments. Here, an agent must actively explore its surroundings to find the answer to a user's question. This task serves as a crucial benchmark for an agent's ability to perceive, navigate, and reason within a physical space.

However, existing EQA paradigms often oversimplify the nature of human-robot interaction by focusing on answering a single, isolated question at a time. In reality, interactions are much more dynamic and complex: several users might pose initial questions and then follow up with subsequent related questions. These questions can arrive asynchronously and possess varying degrees of urgency. For instance, a rescue robot entering a disaster scene may be asked to locate the main staircase, and then receive a high-priority follow-up question about the presence of trapped civilians while exploring. Similarly, a factory assistant may need to simultaneously monitor sensor data, check inventory levels, and respond to urgent safety alerts. Current EQA frameworks are ill-equipped to handle such multifaceted scenarios, limiting their applicability in the real world.

To bridge this gap, we introduce and formulate a novel problem setting: \textbf{E}mbodied \textbf{Q}uestion\textbf{s} \textbf{A}nswering (EQsA). As shown in \Cref{fig:figure_1}, an agent is presented with an initial set of questions and must handle additional follow-up queries that arise later. This setting requires an agent not only to answer individual questions but also to manage a dynamic task list and strategically plan its exploration to gather information for concurrent objectives.

We propose \textbf{Con}current \textbf{E}mbodied \textbf{Q}uestion\textbf{s} \textbf{A}nswering (ConEQsA), a framework tailored for the EQsA problem. The term \emph{concurrent} means that multiple questions are kept in-flight through urgency-aware scheduling and a shared group memory, while physical exploration remains single-threaded. ConEQsA enables concurrent progress by considering the urgency, dependencies, and informational overlap between questions, enabling the agent to operate more efficiently and responsively in complex, dynamic scenarios.

Furthermore, we introduce the \textbf{C}oncurrent \textbf{A}synchronous \textbf{E}mbodied \textbf{Q}uestion\textbf{s} (CAEQs) dataset, the first benchmark designed for EQsA based on the Habitat-Matterport 3D Research Dataset (HM3D) \cite{ramakrishnan2021hm3d}. As a fair evaluation protocol, CAEQs features asynchronous follow-up questions and \emph{human-annotated} urgency labels, and we propose novel metrics including Direct Answer Rate (DAR) and Normalized Urgency-Weighted Latency (NUWL) to evaluate knowledge reuse and urgency-aware responsiveness alongside accuracy.

In summary, our primary contributions are threefold:
\begin{itemize}
    \item We introduce and formulate EQsA, a novel problem setting that requires an agent to answer a set of initial questions and asynchronous follow-up questions concurrently, reflecting realistic human-robot interactions.
    \item We propose ConEQsA, a concurrent agentic framework engineered to solve the EQsA problem, featuring a novel priority planning and question scheduling mechanism.
    \item We develop the CAEQs dataset, the first benchmark for the EQsA task, and introduce a suite of new metrics, including DAR and NUWL, providing a comprehensive and fair evaluation protocol for EQsA methods.
\end{itemize}

\section{Related Work}

\textbf{Embodied Question Answering:} Embodied Question Answering (EQA) was introduced by Das et al.~\cite{das2018embodied} as a task where an agent must navigate a 3D environment to answer a given question, combining navigation, perception, and language understanding. Yu et al.~\cite{yu2019multi} studied multi-target EQA, where the answer may require finding multiple objects, and Tan et al.~\cite{tan2023knowledge} introduced Knowledge-based EQA (K-EQA) that leverages external commonsense knowledge and an episodic scene graph to answer compositional queries. Recent open-vocabulary and foundation-model based efforts (e.g., OpenEQA~\cite{majumdar2024openeqa}) move towards more realistic questions and environments, while Ren et al.~\cite{Ren-RSS-24} improve exploration efficiency for single-question EQA. Beyond the single-question setting, Ginting et al.~\cite{ginting2025mindpalace} propose LA-EQA, emphasizing long-term recall and active exploration over extended horizons, and Zhai et al.~\cite{zhai2025tooleqa} introduce ToolEQA that performs multi-step reasoning via external tools. Zhang et al.~\cite{zhang2025avila} study asynchronous user queries over streaming video, but the agent operates passively on pre-recorded streams rather than actively navigating 3D environments, and does not consider urgency-aware scheduling. CommCP~\cite{zhang2026commcp} further formulates a cooperative multi-agent multi-task EQA problem and coordinates robots via LLM-based communication, but does not study concurrent or asynchronous question scheduling within an episode. In contrast, ConEQsA addresses groups of related questions (EQsA), requiring multi-question planning and persistent reasoning across queries.

\textbf{Memory Mechanisms for Embodied Agents:} Memory plays a critical role in embodied tasks. Classical memory-augmented models (e.g., End-to-End Memory Networks~\cite{sukhbaatar2015end} and Differentiable Neural Computer~\cite{graves2016hybrid}) and recurrent policies for long-horizon navigation~\cite{szot2021habitat} enable agents to store and recall information. In embodied navigation and QA, structured spatial memory has been explicitly studied: Parisotto and Salakhutdinov~\cite{parisotto2017neural} proposed Neural Map, a learned 2D spatial memory for navigation, and Chaplot et al.~\cite{chaplot2020learning} developed Active Neural SLAM to build and use a learned map during exploration. Zhai et al.~\cite{zhai2025memory} introduced a memory-centric EQA architecture (MemoryEQA) that maintains hierarchical memory interacting with all model components, and K-EQA~\cite{tan2023knowledge} similarly highlights episodic scene-graph memory to reduce redundant navigation. More recently, Cai et al.~\cite{cai2025vision2geometry} propose SEER-Bench for sequential embodied reasoning (including sequential EQA) and a 3D spatial-memory-based approach (3DSPMR) that reuses accumulated spatial knowledge across tasks. ConEQsA extends these ideas by enabling \emph{group-level} memory across an entire set of questions, allowing observations gathered for one question to be reused by others and reducing redundancy.

\textbf{Priority-Driven Task Planning:}
Planning under multiple objectives or tasks~\cite{hayes2022practical, ka2024systematic, yang2025preference, liang2025simultaneous, huang2023informable} often requires prioritization. In multi-robot systems, decoupled planning with priority ordering has been studied: for instance, Bennewitz et al.~\cite{bennewitz2001optimizing} optimize priority schemes to minimize joint path lengths, illustrating how priorities influence planning success. In robotic control, hierarchical task stacks enforce high-priority (e.g., safety) tasks before normal tasks~\cite{notomista2023beyond}. Closely related to our setting, Silva and Macharet~\cite{silva2025urgency} study urgency-aware planning and coordination for collaborative LLM agents, highlighting the benefits (and challenges) of explicitly modeling urgency during multi-agent decision making. In reinforcement learning, curricula can be set by difficulty or priority: Cho et al.~\cite{cho2024hard} dynamically schedule harder tasks first to improve learning efficiency. These works motivate the need for task weighting. ConEQsA incorporates weighted prioritization for answering multiple questions and extends prior planning approaches by explicitly integrating question-level importance into EQsA scheduling.

\section{The Embodied Questions Answering Problem}\label{sec:Problem_Formulation}

We formalize a novel, challenging task, Embodied Questions Answering (EQsA), where an embodied agent must schedule and answer a dynamic group of questions within a 3D environment. This contrasts with Embodied Question Answering (EQA), which focuses on a single question.

\textbf{Distribution of Scenarios for EQsA:}
We define a \emph{scenario} as a tuple $\xi := (e, T, g^0, \mathcal{Q}_{init}, \mathcal{Q}_{follow})$, where:
\begin{itemize}
    \item $e$ is a simulated or real 3D scene.
    \item $T$ is the maximum number of time steps allowed for exploration, according to the size of the scenario.
    \item $g^0$ is the agent's initial pose (2D position and orientation at time $t=0$).
    \item $\mathcal{Q}_{init}$ is a set of initial questions given to the agent at the start of the scenario.
    \item $\mathcal{Q}_{follow}$ is a set of follow-up questions that are asynchronously introduced during the exploration process.
\end{itemize}
Each question $q_i$ is a tuple $(s_i, u_i, y_i, t_i)$, where $s_i$ is the question text, $u_i \in (0, 1)$ is its urgency value, $y_i$ is the ground truth answer, and $t_i \geq 0$ is its arrival time. For each question $q_j$ in the initial set $\mathcal{Q}_{init}$, its arrival time is $t_j = 0$. We will use a subscript to indicate the scenario (e.g., $T_{\xi}$) and a superscript for time steps (e.g., $g^t$).

In this work, we consider four-option multiple-choice questions. Thus, the set of possible labels for any question is $\mathcal{Y} := \{\text{'A'}, \text{'B'}, \text{'C'}, \text{'D'}\}$. We assume no prior knowledge of the distribution $\mathcal{D}$, but we can sample a finite dataset of independent and identically distributed scenarios from it.

\textbf{Agent-Environment Interaction:}
In any given scenario $\xi \in \mathcal{D}$, an autonomous agent is initialized at pose $g^0$ with no prior knowledge of the 3D scene $e$. At each time step $t$, the agent navigates to a new pose $g^t$ using a collision-free planner and captures an RGB image $I_c^t \in \mathbb{R}^{H_I \times W_I \times 3}$ along with a depth image $I_d^t$. This visual information is used to construct the agent's memory, which is then leveraged by pre-trained Vision-Language Models (VLMs) to answer active questions.

\textbf{Efficient and Urgency-Aware Questions Scheduling and Exploration:}
In an EQsA task, the agent must take all available questions into account, find an appropriate sequence to answer them, explore the environment to gather necessary information, and provide final answers. Our goal is to optimize the trade-off between overall exploration efficiency (minimizing total time steps) and the latency of individual questions, weighted by their urgency $u_i$.

\section{ConEQsA}\label{sec:Methodology}
In this section, we present the methodology of the ConEQsA framework. First, we provide an overview of its architecture and key components. Subsequently, we elaborate on the priority planning and question scheduling mechanism. Finally, we briefly describe the targeted exploration strategy.

\begin{figure}[t]
    \centering
    \includegraphics[width=\columnwidth]{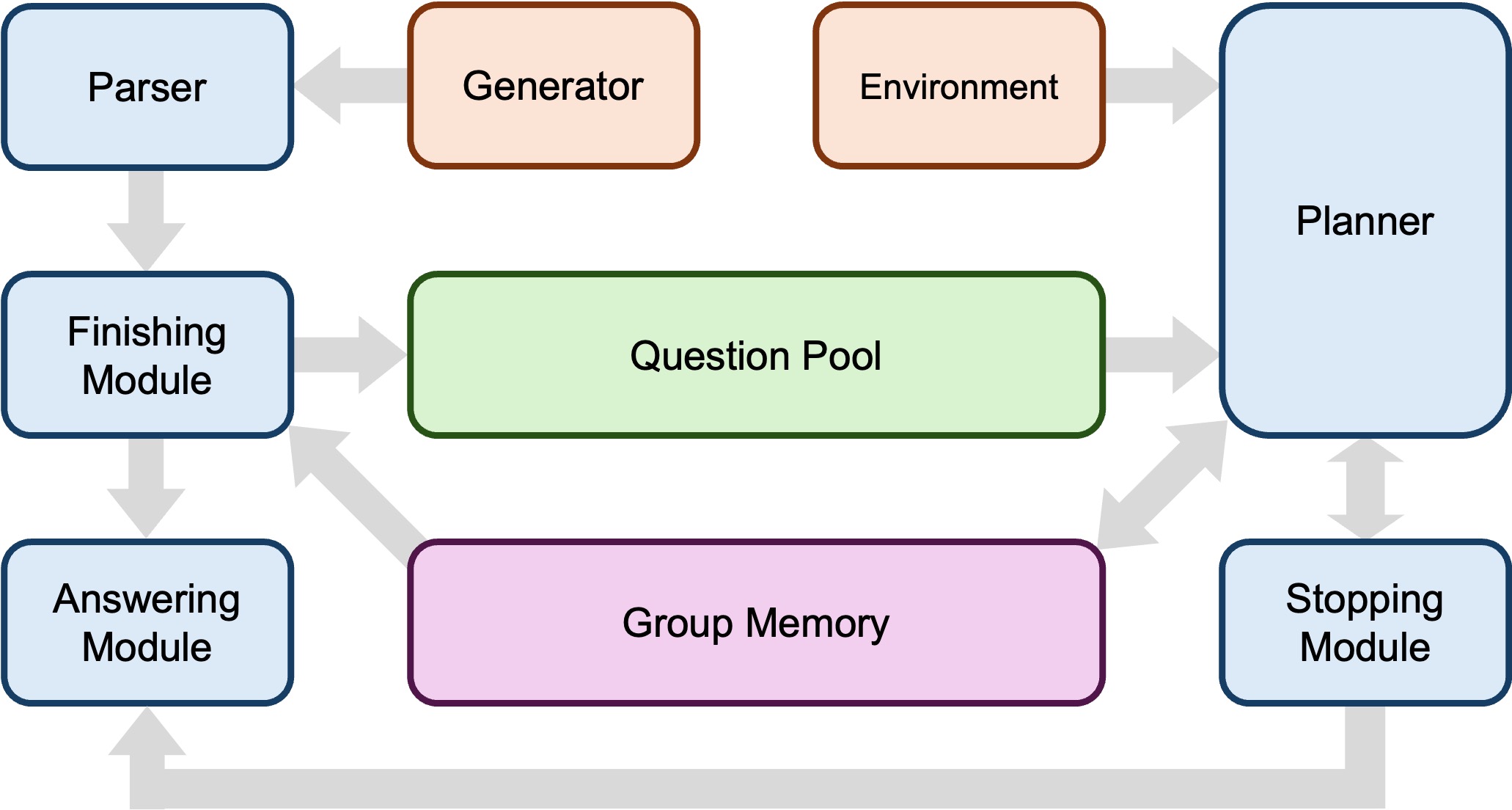}
    \caption{ConEQsA pipeline: parse each incoming question, attempt direct answering from group memory, otherwise enqueue it into a Question Pool; the planner selects a question for question-conditioned exploration, and the answerer responds and updates memory. ``Concurrent'' means multiple questions remain in-flight via scheduling and memory reuse under a single exploring agent.}
    \label{fig:methodology_overview}
\end{figure}

\subsection{Framework Components}
ConEQsA is implemented as distributed microservices communicating via Redis streams. As illustrated in \Cref{fig:methodology_overview}, several specialized modules work asynchronously: Questions undergo semantic parsing and an initial answering attempt. Unsolved questions are then sent to the Question Pool and prioritized. The highest-priority question is selected, explored, and answered.

\textbf{Generator:} The Generator instantiates $\xi$ by resetting the simulator to $(e,g^0)$, then emits a time-stamped question stream: $q_i\in\mathcal{Q}_{init}$ at $t_i=0$ and $q_i\in\mathcal{Q}_{follow}$ according to their arrival times $t_i$. This module operationalizes the task definition and fixes question timings for reproducible evaluation.

\textbf{Parser:} Given $s_i$, the Parser uses an LLM to attach semantic metadata, including urgency $u_i$ and a scope type (local vs.\ global), computed once at arrival.

\textbf{Finishing Module:} The Finishing Module queries Group Memory with $s_i$ and estimates whether the question is answerable with current evidence; answerable questions are sent to the Answering Module, otherwise they are forwarded to the Question Pool.

\textbf{Question Pool:} As illustrated in \Cref{fig:question_pool}, the Question Pool buffers all active questions, maintains dependency status as a DAG, and updates each priority score based on urgency, scope, estimated reward, and dependency.

\textbf{Planner:} The Planner selects the highest-priority question and explores the environment. At each step $t$, it augments the raw observation with a semantic record $\hat{a}^t$ extracted from $(I_c^t,I_d^t,g^t)$: YOLOv11 provides object bounding boxes and category labels; Qwen2.5-VL-7B captions the full frame and cropped object regions to extract attributes; and depth with camera pose is used to approximate each object's 3D position. The fused $\hat{a}^t$ is written to Group Memory.

\textbf{Stopping Module:} During exploration, the Stopping Module periodically retrieves top-$K$ memory items using $s_i$ and the latest $I_c^t$, and uses a VLM-based confidence estimator to decide whether to stop and trigger answering.

\textbf{Answering Module:} The Answering Module produces the final multiple-choice answer by conditioning a VLM on $s_i$ together with the retrieved memory evidence.

\textbf{Group Memory:} Group Memory stores memory items that couple an observation image $I_c^t$ with a structured semantic record $\hat{a}^t$ (e.g., \texttt{room}, \texttt{caption}, \texttt{obj\_bbox}, and \texttt{obj\_pos}). Items are indexed by a CLIP+FAISS retriever for cross-modal
search. When other modules issue a query $s_i$, the top-$K$ returned items are passed into the downstream LLM/VLM prompts.

\subsection{Priority Planning and Questions Scheduling}

\begin{figure}[t]
    \centering
    \includegraphics[width=\columnwidth]{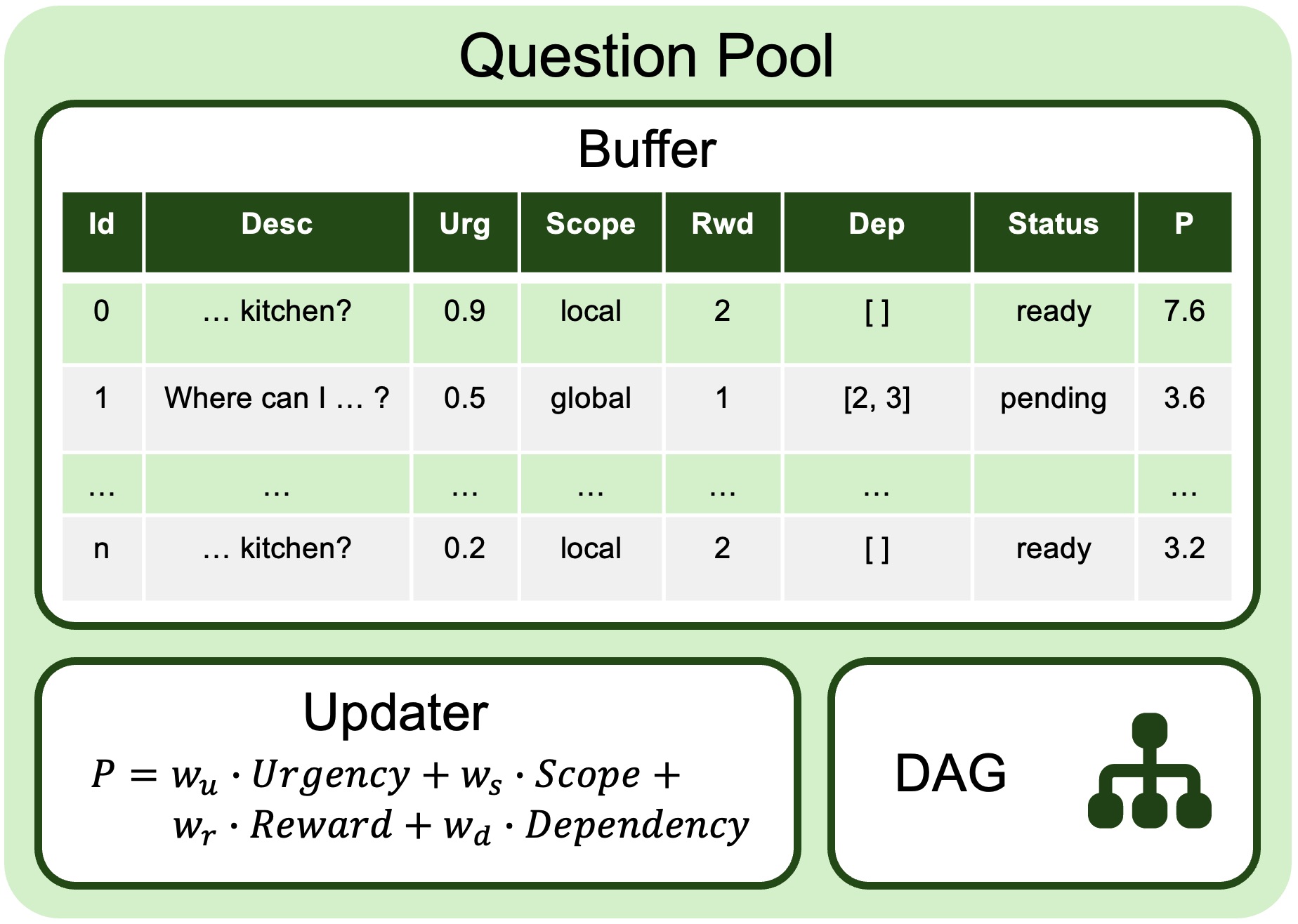}
    \caption{Question Pool maintains buffered questions, resolves cross-question dependencies as a DAG, and assigns each question a priority score based on urgency, scope, estimated reward, and status.}
    \label{fig:question_pool}
\end{figure}

ConEQsA schedules incoming questions through the Question Pool (\Cref{fig:question_pool}). The scheduler aims to reduce urgency-weighted latency by prioritizing (i) urgent questions and (ii) questions whose exploration is likely to produce reusable evidence for other pending questions, rather than processing requests in FIFO order. Formally, we treat each question as a job in an online preemptive scheduling problem whose cost is urgency-weighted latency; $P(q_i)$ approximates the marginal NUWL reduction per unit exploration step for question $q_i$.

When a question arrives or its status changes, the Updater recomputes a priority score for each active question:
\begin{align}
P(q_i) &= w_{u} \cdot Urgency(q_i) + w_{s} \cdot Scope(q_i) \nonumber \\
       &\quad + w_{r} \cdot Reward(q_i) + w_{d} \cdot Dependency(q_i)\,,
\end{align}
where $w_{u}, w_{s}, w_{r}, w_{d}$ are fixed hyperparameters that are selected once on a validation split and shared across all experiments. Each component is defined as follows: 

\begin{itemize}
    \item \textbf{Urgency ($Urgency(q_i)$):} Let $u_i\in[0,1]$ be the urgency inferred from question text at arrival. To emphasize high-urgency questions, we apply a convex transform:
    \begin{equation}
        Urgency(q_i) = -\ln(1 - u_i)\,.
    \end{equation}

    \item \textbf{Scope ($Scope(q_i)$):} The Parser assigns a scope type (local vs.\ global). We prioritize local questions that are typically answerable with nearby observations:
    \begin{equation}
        Scope(q_i) = 
        \begin{cases} 
            1, & \text{if scope is local} \\
            0, & \text{if scope is global}
        \end{cases}
    \end{equation}

    \item \textbf{Reward ($Reward(q_i)$):} To encourage evidence sharing, we estimate how many other unresolved questions may benefit from exploring for $q_i$. Specifically, the Parser extracts a canonical target set $T_i$ (rooms/landmarks and object categories mentioned or implied by $q_i$), and we define $Reward(q_i)$ as the number of other questions $q_j$ in the pool whose targets overlap with $T_i$.

    \item \textbf{Dependency ($Dependency(q_i)$):} Dependencies are predicted by an LLM that links questions into a DAG; an edge $q_j\!\rightarrow\! q_i$ indicates that $q_i$ requires information produced by $q_j$ (e.g., locating an object before checking its state). A question is marked \texttt{ready} iff all prerequisites are resolved; otherwise it is \texttt{pending}:
    \begin{equation}
        Dependency(q_i) = 
        \begin{cases} 
            1, & \text{if status is ready} \\
            0, & \text{if status is pending}
        \end{cases}
    \end{equation}
\end{itemize}

After scoring, the scheduler selects the question with the largest $P(q_i)$ for targeted exploration. Meanwhile, newly arriving questions can still be parsed and may be directly answered by the Finishing Module if sufficient evidence already exists in Group Memory, bypassing exploration.

\subsection{Targeted Exploration}

\begin{figure}[t]
    \centering
    \includegraphics[width=\columnwidth]{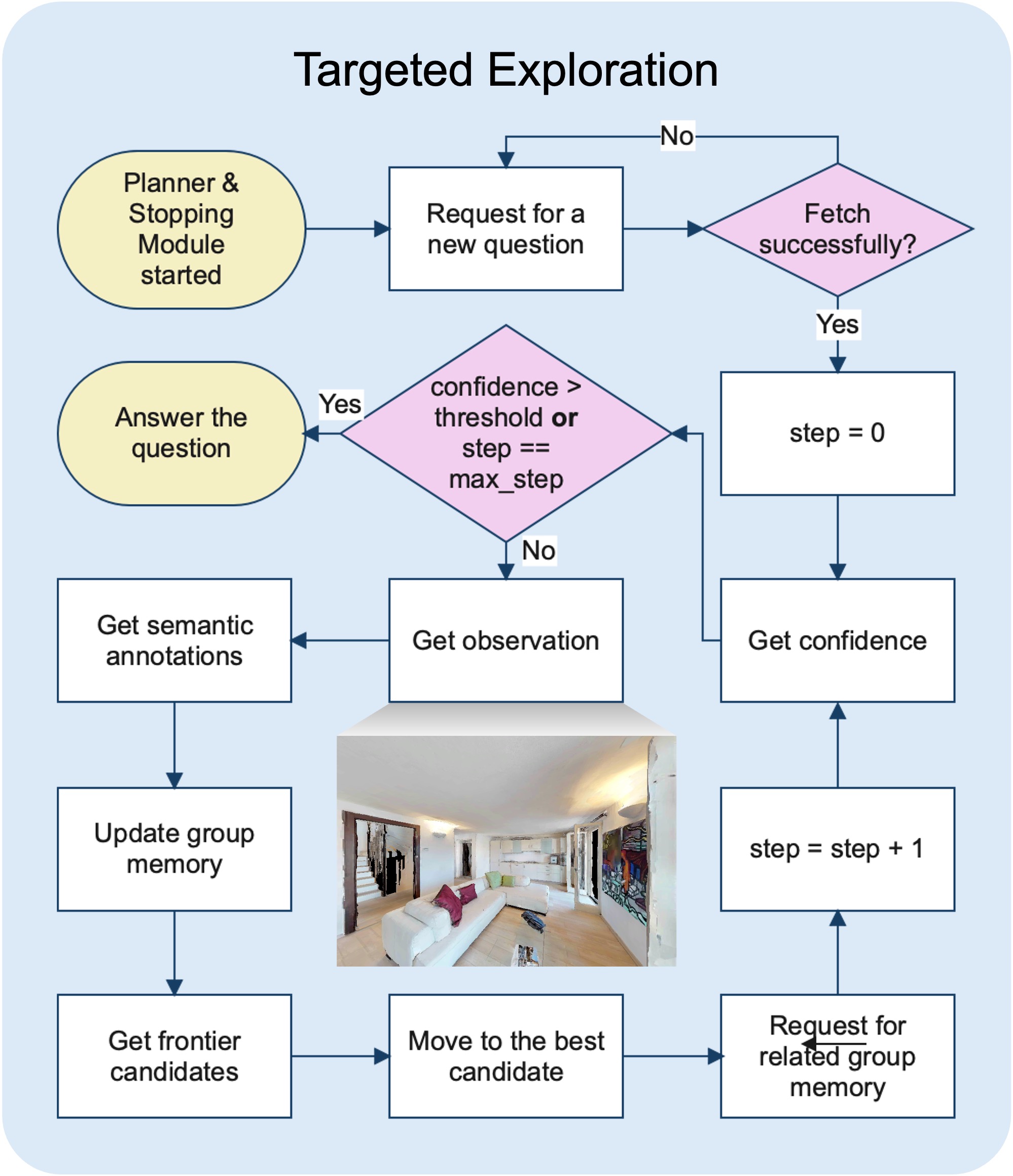}
    \caption{Targeted exploration loop: retrieve relevant memory, update semantic map/memory with new observations, navigate to the next frontier, and stop when the stopping module is satisfied.}
    \label{fig:targeted_exploration}
\end{figure}

To gather evidence for the scheduled question, ConEQsA performs question-conditioned frontier exploration (\Cref{fig:targeted_exploration}), following Explore-EQA~\cite{Ren-RSS-24}. We maintain a 2D semantic map over explored space and score candidate frontiers using question relevance signals. At each step, the current RGB observation is annotated: YOLOv11 detects object instances, and Qwen2.5-VL-7B assigns relevance scores for the question to detected objects and salient regions. These scores are projected onto the semantic map to weight frontiers, guiding the agent toward regions most likely to answer the question; the loop iterates by navigating, updating the semantic map/group memory, and querying the Stopping Module until it is satisfied.

Notably, exploration for a new question $q_i$ starts from the agent's current pose (i.e., the final pose after exploring for $q_{i-1}$). Only for the first question in a scenario does the agent begin at the initial pose $g^0$ specified in the dataset, which is intended to mimic real-world deployment where requests arrive online.

\begin{figure}[t]
    \centering
    \includegraphics[width=\columnwidth]{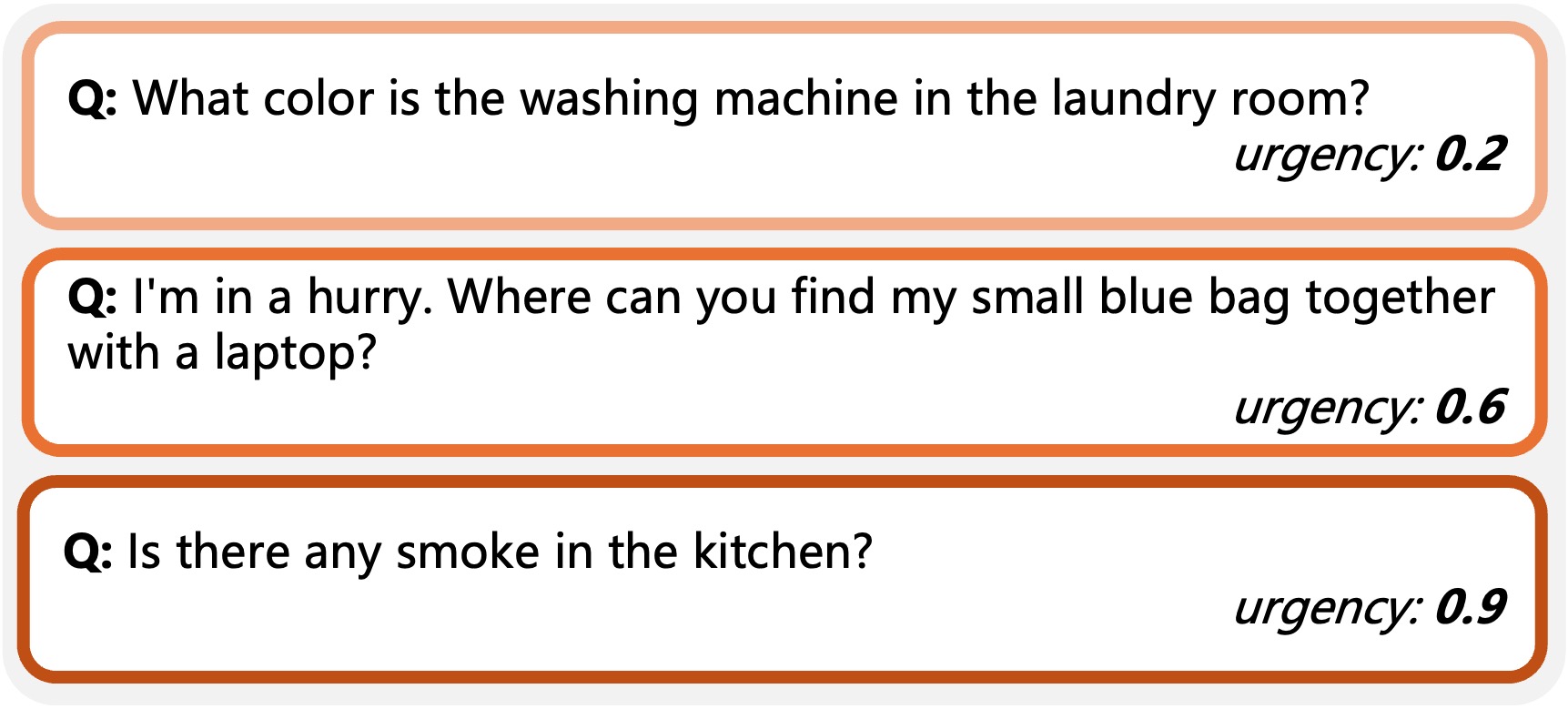}
    \caption{General questions, functional questions, and safety-related questions receive low, medium, and high urgency, respectively.}
    \label{fig:urgency}
\end{figure}

\section{Contributed Benchmark}\label{sec:Benchmark}

\subsection{Concurrent Asynchronous Embodied Questions Dataset}

While existing EQA datasets predominantly focus on scenarios involving a single, static question, we introduce the Concurrent Asynchronous Embodied Questions (CAEQs) dataset to evaluate an agent's ability to schedule, explore, and answer multiple questions that arrive asynchronously and possess varying degrees of urgency.

\begin{figure*}[t]
    \centering
    \includegraphics[width=\textwidth]{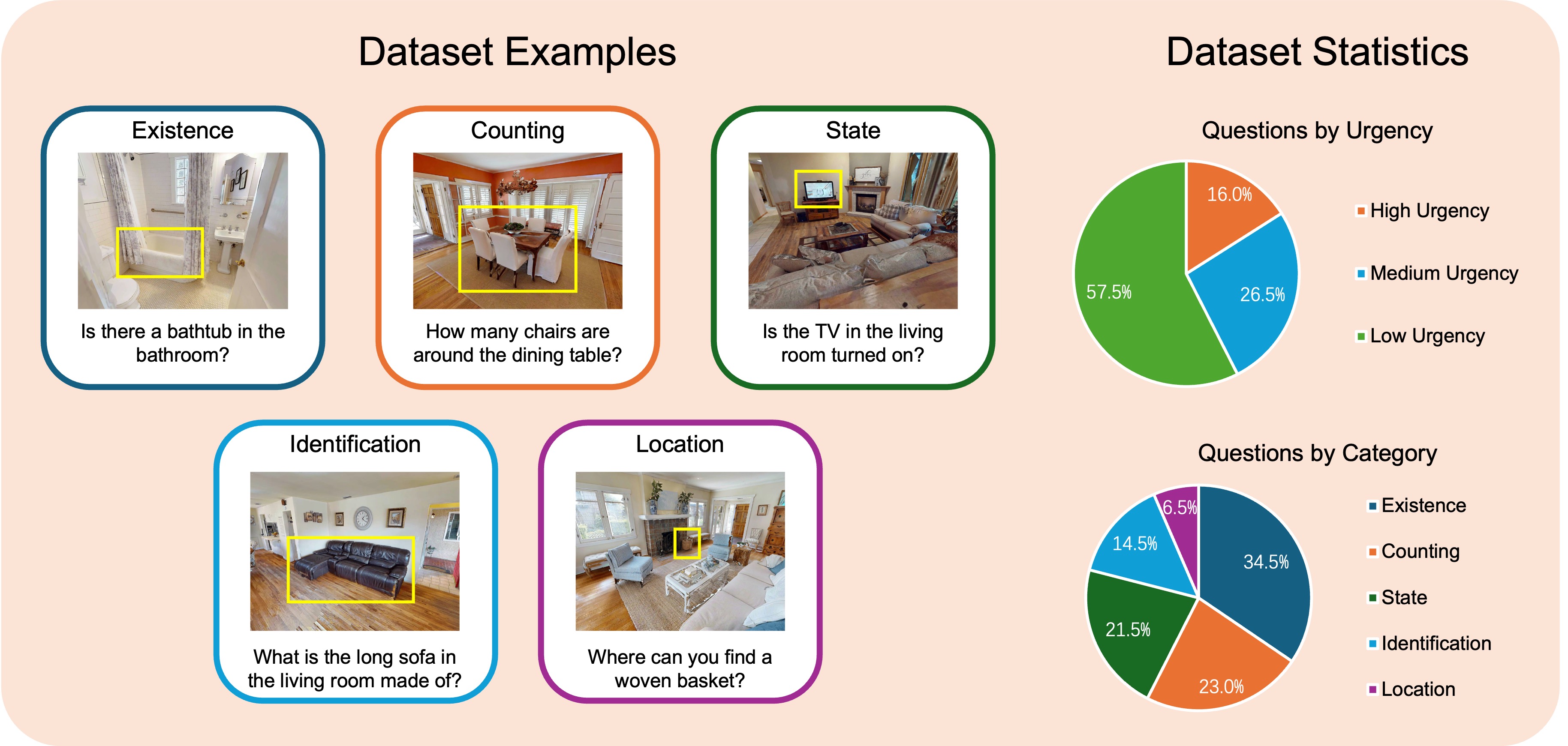}
    \caption{Illustration of the dataset of CAEQs benchmark. Left: dataset examples, including 5 categories of questions. Right: dataset statistics, detailing the distribution across categories and \emph{human-annotated} urgency levels.}
    \label{fig:dataset}
\end{figure*}

As shown in \Cref{fig:dataset}, the CAEQs dataset is constructed using 40 high-quality, photo-realistic residential scenes selected from the Habitat-Matterport 3D (HM3D) \cite{ramakrishnan2021hm3d} Research Dataset. For each scene, we prepare a set of 5 questions, which are partitioned into two types: 3 initial questions presented at the start of a scenario, and 2 follow-up questions, each with a 120-second delay after the previous question. To ensure a fair and unbiased evaluation, the five questions for each scene are randomly assigned to the initial or follow-up category. All 200 questions in the CAEQs dataset were manually authored and refined through multiple rounds of review to ensure clarity, relevance, and correctness. The corresponding ground-truth answers were also meticulously verified.

In the CAEQs dataset, each question is associated with a \emph{manually annotated} urgency score according to its semantic content. They are categorized into three levels: low ($[0, 0.3)$), medium ($[0.3, 0.7)$), and high ($[0.7, 1.0]$), which constitute 57.5\%, 26.5\%, and 16.0\% of the dataset, respectively. As shown in \Cref{fig:urgency}, questions pertaining to general information, functionality, and safety are assigned low, medium, and high urgency, respectively. We follow this rubric to assign consistent urgency values across the dataset.

It is worth mentioning that the urgency scores in the dataset are only for evaluation purposes. The agent has no prior knowledge of these scores and must infer the urgency of each question based solely on its content.

The 200 questions in the CAEQs dataset are distributed across five semantic categories, reflecting a variety of tasks an embodied agent might encounter:
\begin{itemize}
    \item \textbf{Existence (34.5\%):} Asking if an object is present, e.g., \textit{"Is there a bathtub in the bathroom?"}
    \item \textbf{Counting (23.0\%):} Asking for the number of objects, e.g., \textit{"How many chairs are around the dining table?"}
    \item \textbf{State (21.5\%):} Asking about the state of an object, e.g., \textit{"Is the refrigerator door open?"}
    \item \textbf{Location (12.0\%):} Asking where an object is located, e.g., \textit{"Where is the vacuum cleaner?"}
    \item \textbf{Relation (9.0\%):} Asking about the relation between objects, e.g., \textit{"What is on top of the coffee table?"}
\end{itemize}

Our formulation \Cref{sec:Problem_Formulation} requires every question to have four multiple-choice answers. For consistency, if a question starts with fewer than four options, we add dummy choices like "D) (Do not choose this option)".

\subsection{Evaluation Metrics}
To comprehensively evaluate the agent's performance in the EQsA task, we introduce a set of metrics that measure not only the correctness of the answers but also the efficiency and responsiveness of the agent's scheduling and exploration strategy. Let $\mathcal{Q}$ be the set of all questions in a given scenario, where each question $q_i \in \mathcal{Q}$ has an urgency value $u_i$.

\subsubsection{Accuracy (Acc)}
The accuracy is calculated as:
\begin{equation}
\text{Acc} = \frac{1}{|\mathcal{Q}|} \sum_{q_i \in \mathcal{Q}} \mathbb{I}(\hat{y}_i = y_i)
\end{equation}
where $\hat{y}_i$ is the agent's predicted answer for question $q_i$, $y_i$ is the ground truth answer, and $\mathbb{I}(\cdot)$ is the indicator function, which is 1 if the condition is true and 0 otherwise.

\subsubsection{Direct Answer Rate (DAR)}
This metric quantifies the agent's ability to leverage its memory to answer questions without performing any physical exploration. A higher DAR suggests a more effective use of accumulated knowledge. It is calculated as:
\begin{equation}
\text{DAR} = \frac{|\mathcal{Q}_{direct}|}{|\mathcal{Q}|}
\end{equation}
where $\mathcal{Q}_{direct}$ is the subset of questions that the agent answers from memory (i.e., with zero exploration steps).

\subsubsection{Normalized Steps (NS)}
For a given question $q_i$, $\text{ns}_i$ is defined as the ratio of $\text{used\_steps}_i$ to $\text{max\_steps}_i$, where $\text{used\_steps}_i$ is the number of exploration steps taken for question $q_i$, and $\text{max\_steps}_i$ is the maximum step allocation for that question. A lower NS value indicates higher efficiency, and $\text{ns}_i = 0$ indicates $q_i$ is answered directly without exploration. The average value of all questions is then,
\begin{equation}
\text{NS} = \frac{1}{|\mathcal{Q}|} \sum_{q_i \in \mathcal{Q}} \frac{\text{used\_steps}_i}{\text{max\_steps}_i}
\end{equation}

\subsubsection{Normalized Urgency-Weighted Latency (NUWL)}
This is a key metric designed to measure the agent's ability to schedule questions effectively. It captures the total latency experienced by all questions, weighted by their respective urgency values. A lower NUWL score is desirable, as it indicates that high-urgency questions were answered with minimal delay. The NUWL is defined as:
\begin{equation}
\text{NUWL} = \frac{1}{|\mathcal{Q}|} \sum_{q_i \in \mathcal{Q}} u_i \cdot \left( \sum_{q_j \in \mathcal{Q}, t_j^{start} < t_i^{req}} \text{ns}_j + \text{ns}_i \right)
\end{equation}
where $u_i$ is the urgency of question $q_i$, $\text{ns}_i$ is its normalized steps, $t_i^{req}$ is the time when $q_i$ was requested, and $t_j^{start}$ is the time when question $q_j$ is selected and processed.

\section{Experimental Evaluation}\label{sec:Experimental_Evaluation}

In this section, we conduct a series of experiments to validate the effectiveness of our proposed ConEQsA framework. Our evaluation is designed to answer two primary research questions:
\begin{enumerate}
    \item How important is the group memory shared among a set of questions, and to what extent does a concurrent problem-solving methodology outperform a sequential one?
    \item How necessary is priority planning, and what is the relative importance of its key components?
\end{enumerate}
To address these questions, we perform a main experiment comparing ConEQsA against two baselines and a comprehensive ablation study to dissect the contributions of our priority planning mechanism.

\subsection{Implementation Details}
To balance performance and cost, we utilize GPT5-mini \cite{openai2025_gpt5_system_card} as the main VLM in most modules, and an effective LLM GPT-OSS-20B \cite{agarwal2025gpt} for non-vision tasks. Semantic annotation tasks are accomplished by a YOLOv11 \cite{yolo11_ultralytics} model, and a lightweight VLM Qwen2.5-VL-7B \cite{Qwen2.5-VL}. For the simulated experiments, we run the selected scenarios from the HM3D dataset \cite{ramakrishnan2021hm3d} in the Habitat simulator on a Linux machine with an NVIDIA RTX 3080 Ti GPU. 

In our current implementation, the average wall-clock LLM/VLM inference latency to answer a single question is around 5 minutes; by issuing multiple independent API calls in parallel, this latency can be reduced to under 3 minutes, which is comparable to prior EQA agent systems.

\subsection{Baselines}
We compare ConEQsA against two strong baselines adapted for the multi-question EQsA task:
\begin{itemize}
    \item \textbf{Explore-EQA:} This baseline \cite{Ren-RSS-24} handles questions sequentially and has no group memory shared among questions. This is a strong and widely-used baseline in recent EQA works.
    \item \textbf{Memory-EQA:} This baseline \cite{zhai2025memory} has achieved state-of-the-art (SOTA) accuracy in EQA tasks, which also processes questions sequentially. We adapt it to our EQsA setting by using its hierarchical memory module as our group memory, and employing the same frontier-based exploration policy for each question.
\end{itemize}
All methods are evaluated on the CAEQs dataset, and we report the average performance across all 40 scenarios using the metrics defined in \Cref{sec:Benchmark}.

\subsection{Main Results: ConEQsA vs. Sequential Baselines}

\begin{table}[t]
\centering
\caption{Main experiment results comparing ConEQsA with strong sequential baselines. Our method demonstrates both greater performance and responsiveness.}
\label{tab:main_experiment}
\begin{adjustbox}{width=0.95\columnwidth,center}
\begin{tabular}{@{}lcccc@{}}
\toprule
\textbf{Method} & \textbf{Acc} $\uparrow$ & \textbf{DAR} $\uparrow$ & \textbf{NS} $\downarrow$ & \textbf{NUWL} $\downarrow$ \\
\midrule
ConEQsA (Ours) & \textbf{0.65} & \textbf{0.09} & \textbf{0.321} & \textbf{0.204} \\
Explore-EQA \cite{Ren-RSS-24} & 0.62 & 0.00 & 0.472 & 0.551 \\
Memory-EQA \cite{zhai2025memory} & 0.64 & 0.00 & 0.410 & 0.474 \\
\bottomrule
\end{tabular}
\end{adjustbox}
\end{table}

The results, summarized in \Cref{tab:main_experiment}, clearly demonstrate the comprehensive advantages of our proposed ConEQsA framework over sequential baselines. Our method shows substantial gains in efficiency and responsiveness, while achieving a comparable accuracy to Memory-EQA, the current SOTA in EQA tasks.

\subsubsection{Superior Efficiency and Responsiveness}
The most striking improvements are observed in Normalized Steps (NS) and Normalized Urgency-Weighted Latency (NUWL). ConEQsA reduces the NS to 0.321, a significant improvement over both Memory-EQA (0.410) and Explore-EQA (0.472). This highlights the effectiveness of the concurrent processing architecture. By maintaining a shared group memory, ConEQsA can often gather information pertinent to multiple questions during a single exploration trajectory, thus minimizing redundant navigation.

This efficiency directly translates into superior responsiveness, as measured by the NUWL metric. Our method achieves an NUWL of 0.204, which is a 57\% reduction compared to Memory-EQA (0.474) and a 63\% reduction compared to Explore-EQA (0.551). This dramatic improvement is a clear indicator that our priority planning mechanism (\Cref{sec:Methodology}) successfully identifies and expedites high-urgency questions, preventing them from being delayed by less critical queries. The sequential nature of the baselines makes them inherently incapable of such dynamic re-prioritization, especially when they encounter follow-up questions, causing a linear accumulation of latency, which is heavily penalized by the NUWL metric.

\subsubsection{Effective Knowledge Reuse}
A key feature of our framework is the ability to answer questions without any exploration, quantified by the Direct Answer Rate (DAR). ConEQsA achieves a DAR of 9.0\%, while both baselines score 0\%. This is because the baselines, by design, tackle each question with a new, isolated exploration phase. In contrast, our Finishing Module actively queries the shared group memory before initiating exploration. As the agent navigates the environment to answer one question, it continuously populates the memory with observations. This accumulated knowledge can then be sufficient to immediately answer subsequent or newly-arrived questions, particularly those whose scope falls within already-explored areas. This result strongly supports our problem formulation's emphasis on treating a group of questions holistically rather than as a series of independent tasks.

In summary, the main experiment results robustly validate our approach. By shifting from a sequential to a concurrent and asynchronous paradigm with shared memory and intelligent priority scheduling, ConEQsA demonstrates marked superiority in handling the complex, dynamic, and urgency-aware challenges presented by the CAEQs benchmark.

\subsection{Ablation Study: Priority Planning Components}

\subsubsection{Experimental Setup}
To investigate the contribution of each component in our priority planning mechanism (\Cref{sec:Methodology}), we conduct an ablation study where we systematically disable each factor from the priority score calculation:
\begin{itemize}
    \item \textbf{w/o Priority:} Disables the entire priority planning module, reverting to a FIFO queue, while the group memory remains active.
    \item \textbf{w/o Urgency:} Removes the urgency component from the priority calculation.
    \item \textbf{w/o Scope:} Removes the scope component that distinguishes between local and global questions.
    \item \textbf{w/o Reward:} Removes the reward component that encourages the agent to explore places with multiple related questions.
    \item \textbf{w/o Dependency:} Removes the dependency-checking component.
\end{itemize}

\begin{table}[t]
\centering
\caption{Ablation study on the components of the priority planning mechanism. The results confirm that each component contributes to the overall performance, with ConEQsA producing the best performance when all components are intact.}
\label{tab:ablation_study}
\begin{adjustbox}{width=0.95\columnwidth,center}
\begin{tabular}{@{}lcccc@{}}
\toprule
\textbf{Method} & \textbf{Acc} $\uparrow$ & \textbf{DAR} $\uparrow$ & \textbf{NS} $\downarrow$ & \textbf{NUWL} $\downarrow$ \\
\midrule
ConEQsA & \textbf{0.65} & \textbf{0.090} & \textbf{0.321} & \textbf{0.204} \\
\midrule
w/o Priority & 0.63 & 0.085 & 0.398 & 0.232 \\
w/o Urgency & 0.63 & 0.085 & 0.385 & 0.223 \\
w/o Scope & \textbf{0.65} & \textbf{0.090} & 0.336 & 0.213 \\
w/o Reward & 0.64 & 0.085 & 0.374 & 0.226 \\
w/o Dependency & 0.64 & 0.085 & 0.341 & 0.218 \\
\bottomrule
\end{tabular}
\end{adjustbox}
\end{table}

\subsubsection{Results and Analysis}
The results of the ablation study are presented in \Cref{tab:ablation_study}. As shown, removing any component of the priority planning system leads to a degradation in performance, particularly in Normalized Steps (NS) and Normalized Urgency-Weighted Latency (NUWL), confirming that each element plays a vital role. Disabling priority planning (w/o Priority) results in a significant performance drop compared to the full model, increasing NUWL by 13.7\% and NS by 24.0\%. This highlights the fundamental importance of intelligent scheduling over a FIFO approach. 

The priority planning strategy comprises four components: \textit{Urgency}, \textit{Scope}, \textit{Reward}, and \textit{Dependency}. Removing \textit{Urgency} leads to a notable increase in NUWL, confirming its importance in ensuring timely responses to critical questions. \textit{Reward} and \textit{Dependency} also prove valuable, as their removal increases both NS and NUWL. \textit{Scope} has a smaller but still noticeable impact, suggesting that prioritizing locally-scoped questions is a useful approach.

Overall, the ablation study validates our design that a multi-faceted priority planning mechanism is crucial for achieving efficient and urgency-aware performance in the complex EQsA task.

\section{Discussions}\label{sec:Discussions}

In this paper, we formulate the Embodied Questions Answering (EQsA) problem, propose the ConEQsA framework, and build the CAEQs benchmark. We hope these can serve as a standardized testbed for future multi-question EQA methods.

With continued progress in model distillation and hardware acceleration, the latency of LLM/VLM inference is expected to further decrease. Moreover, such inference can often be overlapped with the agent's movements and other system-level execution. In contrast, physical interaction obeys non-negotiable constraints: each real-world action step introduces unavoidable time delay, energy consumption, and mechanical wear. Therefore, reducing the number of environment interaction steps is fundamental to improving real-world responsiveness. This is precisely what NUWL measures and what increasing DAR aims to promote, which jointly reflect the core value of ConEQsA.

Despite improved concurrent scheduling, ConEQsA is still constrained by the current accuracy and robustness of single-question EQA components (e.g., perception, navigation, and answer correctness). As a result, supporting multiple questions in EQsA does not yet translate into a practically useful system in the wild. We therefore primarily position ConEQsA and the CAEQs benchmark as a fair evaluation methodology for EQsA, intended to facilitate systematic comparison and future progress as underlying EQA accuracy improves.

A major avenue for future effort lies in extending our framework to not only be \emph{multi-question}, but also \emph{multi-agent}. Under this setup, multiple agents collaboratively explore the environment to solve the EQsA task, potentially leading to even greater efficiency and responsiveness. This would introduce new challenges in decentralized coordination and task allocation.

\bibliographystyle{ieeetr} 
\bibliography{bib}
\end{document}